# Graph-Based Self-Healing Tool Routing
# for Cost-Efficient LLM Agents

Neeraj Bholani
*February 2026*

Abstract

Tool-using LLM agents face a reliability–cost tradeoff: routing every decision through the LLM improves correctness but incurs high latency and inference cost, while pre-coded workflow graphs reduce cost but become brittle under unanticipated compound tool failures. We present Self-Healing Router, a fault-tolerant orchestration architecture that treats most agent control-flow decisions as routing rather than reasoning. The system combines (i) parallel health monitors that assign priority scores to runtime conditions such as tool outages and risk signals, and (ii) a cost-weighted tool graph where Dijkstra's algorithm performs deterministic shortest-path routing. When a tool fails mid-execution, its edges are reweighted to infinity and the path is recomputed — yielding automatic recovery without invoking the LLM. The LLM is reserved exclusively for cases where no feasible path exists, enabling goal demotion or escalation. Prior graph-based tool-use systems (ControlLLM, ToolNet, NaviAgent) focus on tool selection and planning; our contribution is runtime fault tolerance with deterministic recovery and binary observability — every failure is either a logged reroute or an explicit escalation, never a silent skip. Across 19 scenarios spanning three graph topologies (linear pipeline, dependency DAG, parallel fan-out), Self-Healing Router matches ReAct's correctness while reducing control-plane LLM calls by 93% (9 vs 123 aggregate) and eliminating the silent-failure cases observed in a well-engineered static workflow baseline under compound failures.

## Key Concepts

This paper combines two engineering ideas. To help orient readers:

Parallel Health Monitors are lightweight modules — intent classifiers, risk detectors, tool health checkers — that run in parallel on every request. Each produces a priority score. The highest-priority signal (e.g., a tool failure at 0.99 outbids a routine intent at 0.5) determines what the system pays attention to next. This replaces the expensive LLM as the decision-maker for routine situations.

Tool Graph with Dijkstra Routing is the core routing engine. Tools are nodes, connections are edges with cost weights, and Dijkstra's shortest-path algorithm finds the cheapest working route. When a tool goes down, its edges get infinite weight and the engine automatically finds an alternative path. No human (or LLM) needs to intervene.

Together: the Health Monitors decide what matters right now; the Tool Graph calculates the best path given those priorities. The LLM is only called when the graph cannot find any path to the goal — which, in our benchmarks, means 93% fewer LLM calls than a pure-LLM approach.

## Contributions

1. Deterministic fault-tolerant routing for tool orchestration. We define a runtime recovery mechanism that reweights failed tool edges to infinity and recomputes shortest paths via Dijkstra, making recovery complexity invariant to the number of simultaneous failures — one graph recomputation regardless of K concurrent tool outages.

2. Separation of concerns: monitors → routing → LLM as last resort. Parallel health monitors produce priority signals; graph routing handles routine substitutions; the LLM is invoked only when the graph returns no path, for goal demotion or escalation. This reduces control-plane LLM calls by 93% in our benchmarks.

3. Binary observability guarantee for recovery outcomes. Unlike static workflow graphs where compound failure states can produce partial completion without a domain-level error signal, the router returns either a valid rerouted path or null — enabling explicit escalation when alternatives are exhausted.

4. Cross-topology evaluation exposing a practical gap in well-engineered static workflows. Across 19 scenarios spanning three topologies, Self-Healing Router achieves 19/19 correctness with 9 LLM reasoning calls, compared to ReAct's 123 calls and a static workflow baseline that exhibits 3 silent failures under compound fault conditions.

## 1. The Problem: LLM-Per-Decision vs. Pre-Coded Brittleness

Every production LLM agent today chooses between two extremes. The first approach — exemplified by ReAct, and adopted by production SDKs like OpenAI Agents SDK and Claude Agent SDK — routes every tool selection, error recovery, and control flow decision through the LLM. This ensures correctness but incurs O(n) LLM calls per task, where n is the number of decisions. Our benchmark shows ReAct requiring 123 total LLM reasoning calls across 19 scenarios (mean 6.5 calls per scenario, ranging from 3 for simple happy-path tasks to 15 for complex multi-failure recovery), because each Thought-Action-Observation cycle in ReAct requires a separate LLM inference.

The second approach — exemplified by LangGraph and similar workflow frameworks — defines transitions as programmatic conditional edges (since LangGraph v1.0, nodes can dynamically select their next destination via the Command() API), eliminating LLM overhead. A well-engineered state machine can handle any single anticipated failure by adding explicit fallback edges. But the number of edges required grows combinatorially: for N tools with K possible failure modes each, covering all simultaneous failure combinations requires pre-coded transitions for each of the K^N possible failure combinations — an exponential explosion in N. In practice, teams code the common single-failure cases and hope multi-failure scenarios are rare. When unanticipated combinations occur, the state machine silently proceeds — producing incorrect results without any error signal. A workflow engine can mitigate this by adding explicit end-to-end completion assertions, but such assertions must enumerate the compound-failure cases they check against; this either remains incomplete or grows combinatorially with the failure space.

A third approach has emerged in production: agent SDKs from OpenAI and Anthropic. These wrap LLM reasoning in ergonomic tooling — structured error handling, retry logic, guardrails — but fundamentally inherit the same LLM-per-decision cost. When a tool fails, the error is passed back to the LLM for reasoning about what to do next. In practice, their error-handling loops often resemble ReAct-style patterns where each tool failure routes back to the LLM for reasoning.

The false dichotomy is this: all three approaches assume decisions must be either LLM-generated (expensive, correct) or pre-coded (cheap, brittle). But most agent decisions are neither novel nor pre-anticipated — they are structurally routine but contextually variable. A refund can go through Stripe or Razorpay; a notification can go via email or SMS. The decision logic is "find the cheapest working path" — which is a graph algorithm, not a reasoning task. And critically, graph pathfinding handles all failure combinations with the same algorithm call, regardless of how many tools are simultaneously down.

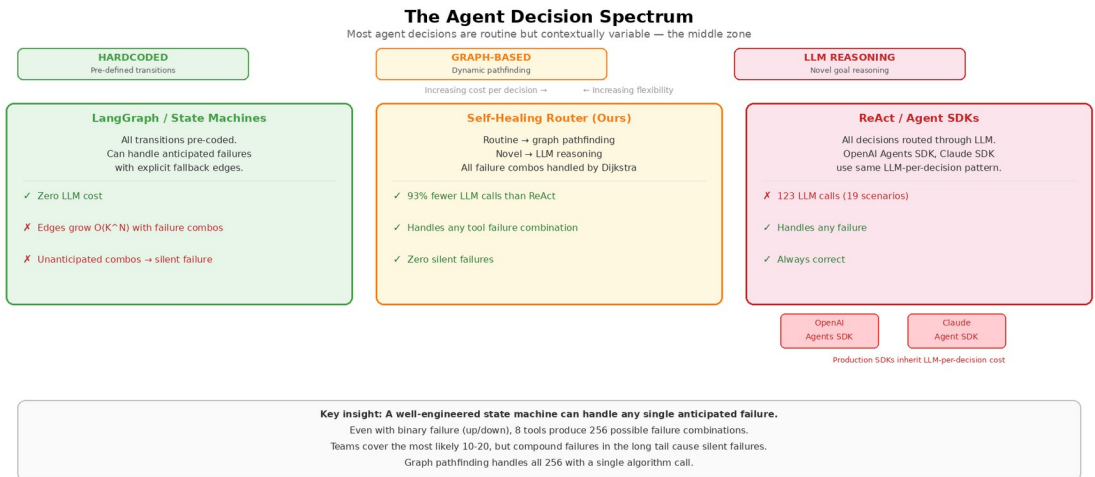

*Figure 1: The agent decision spectrum. LangGraph operates at the hardcoded extreme; production Agent SDKs (OpenAI, Claude) at the LLM extreme. Self-Healing Router occupies the middle — graph algorithms for routine decisions, LLM only for novel reasoning. Key: failure combinations grow exponentially for state machines but are handled by a single Dijkstra call.*

## 2. Architecture: Health Monitors + Tool Graph

Our architecture has three layers that separate attention (what matters now), action planning (what tools to use), and reasoning (what to do when plans fail).

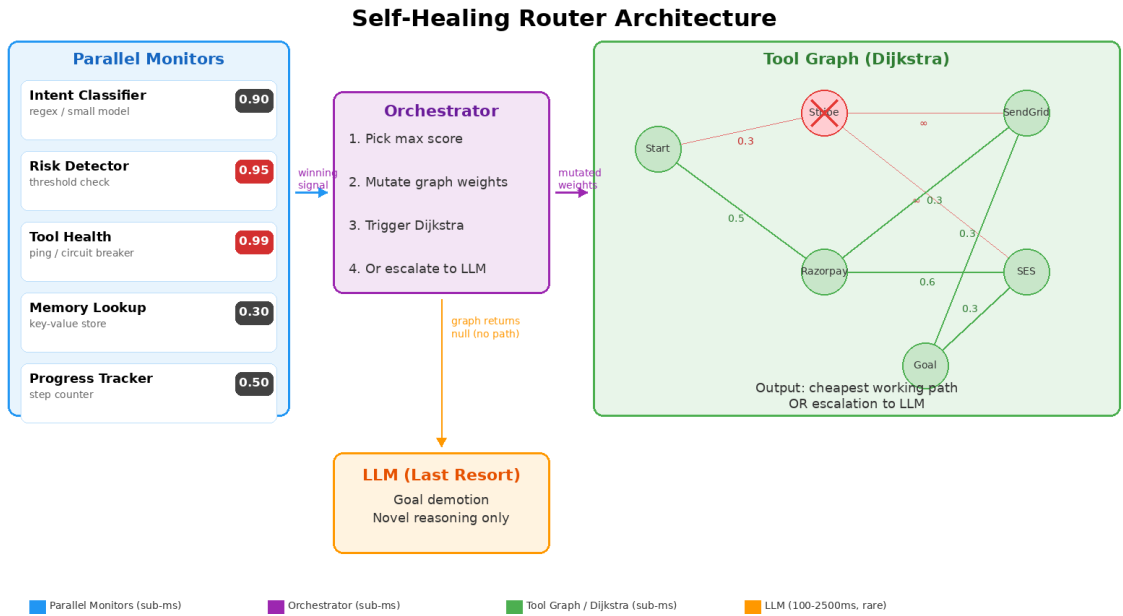

*Figure 2: Self-Healing Router Architecture. Parallel health monitors compete on priority; the tool graph handles routing; the LLM is invoked only for goal demotion.*

### 2.1 Parallel Health Monitors: Priority Competition

The architecture runs cheap parallel health monitors — intent classifiers, risk detectors, tool health monitors, memory lookups, and progress trackers — that each produce priority-scored signals. Signals compete: a risk detector's 0.95-priority signal outbids an intent classifier's 0.5 signal, redirecting the system's attention to the safety concern. This replaces the LLM's role as attention allocator with deterministic competition that costs microseconds instead of hundreds of milliseconds.

Each module is a lightweight function — regex matching for intent classification, threshold comparison for risk detection, ping-based health checks for tool monitoring. None require LLM inference. The priority competition itself is a simple max() operation over float scores. The winning signal is broadcast to the orchestrator, which uses it to decide the next action. Figure 3 illustrates this with a concrete example: when a $15,000 refund request arrives, the risk detector's priority (0.95) outbids the intent classifier's (0.90), causing the orchestrator to escalate rather than process the refund — all without an LLM call.

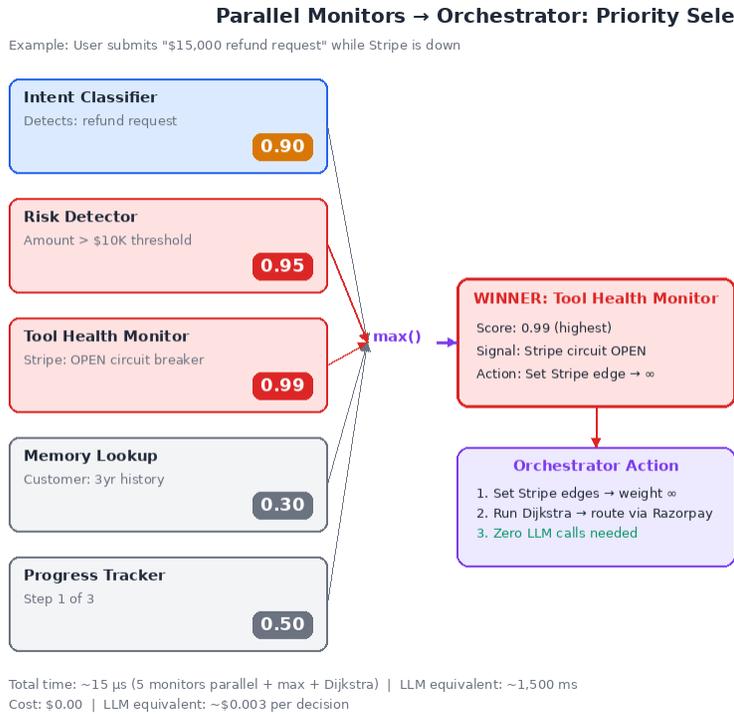

*Figure 3: Priority selection in action. Five parallel monitors score the incoming request. The Tool Health Monitor wins with 0.99 (Stripe circuit breaker open), and the orchestrator sets Stripe edges to infinity before running Dijkstra — all without LLM involvement.*

## 2.2 Tool Graph: Dynamic Cost-Weighted Routing

The tool graph encodes tool capabilities as cost-weighted nodes with preconditions and dependency edges. Given a goal and the current system state (including which tools are healthy), Dijkstra's algorithm (Dijkstra, 1959) finds the cheapest valid path. When a tool fails mid-execution, the graph updates: the failed node's edges receive infinite cost, and Dijkstra finds the next best path automatically. This handles the entire class of "routine but contextually variable" decisions without any LLM involvement.

Figure 4 shows this concretely. In normal operation (panel A), Dijkstra finds the optimal path START to CRM to Stripe to Email to GOAL with total cost 4.0. When Stripe and Email fail simultaneously (panel B), their edge weights go to infinity. Dijkstra re-runs and finds the backup path through Razorpay and SMS with cost 6.0. Backup paths are always present in the graph; they simply cost more. Dijkstra never selects them while cheaper primaries are healthy, but the moment a primary fails, the backup becomes optimal. No new edges need to be added.

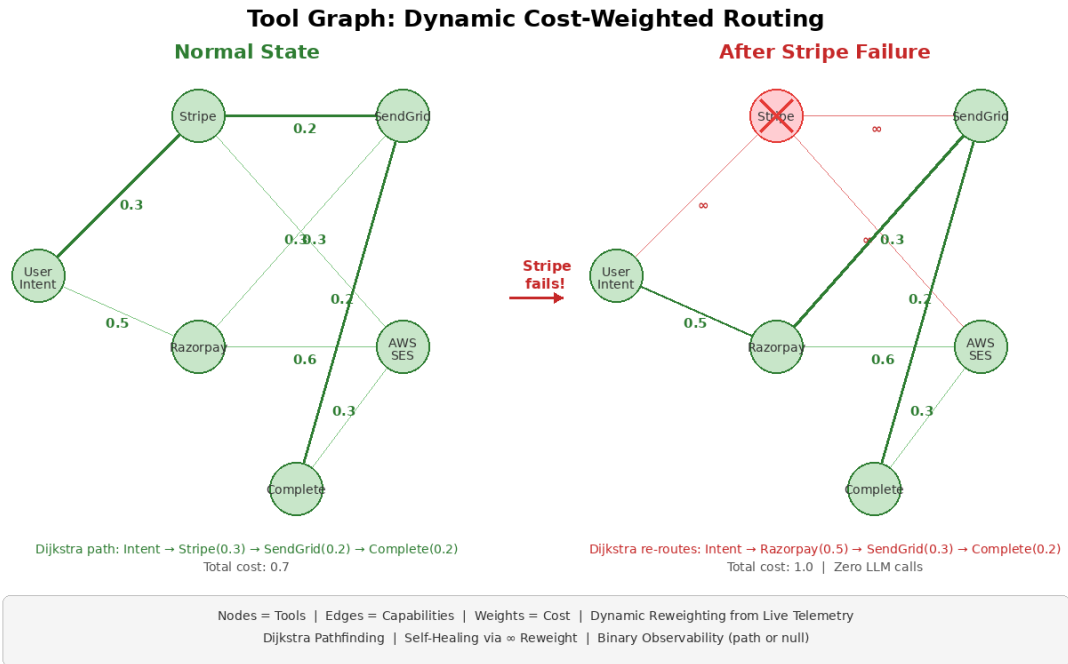

*Figure 4: Dynamic tool graph. (A) Normal: Dijkstra finds cheapest path through primary tools. (B) After Stripe and Email fail: edge weights go to infinity, Dijkstra reroutes through Razorpay and SMS. Backup paths are always present; they simply cost more.*

## 2.3 Path Recalculation: The Algorithm

The path recalculation mechanism is the core technical contribution. When a tool fails during execution, the following sequence occurs in deterministic order:

Step 1 — Failure Detection: The Tool Health Monitor detects the failure (HTTP timeout, error response, or connection refused) and fires a high-priority signal (0.99) to the orchestrator.

Step 2 — Graph Update: The orchestrator receives the signal and sets the weight of all edges touching the failed node to infinity (Float.MAX_VALUE). The graph structure is unchanged — only edge weights are modified. This is O(degree) where degree is the number of edges connected to the failed node, typically 2-4.

Step 3 — Dijkstra Re-run: The orchestrator re-invokes Dijkstra's shortest-path algorithm from the current execution point to the goal node. Because failed edges now have infinite weight, Dijkstra naturally avoids them and finds the cheapest remaining path. This is $O((V + E) \log V)$ with a binary heap, where V is the number of tool nodes and E is the number of edges — sub-millisecond on graphs of this size and negligible compared to tool and network latency (typically 50-2000 ms per API call).

Step 4 — Execution Continues: The new path is returned. The orchestrator resumes execution at the next node in the new path, skipping any work already completed. If no finite-cost path exists (all alternatives exhausted), Dijkstra returns null and the orchestrator escalates to the LLM for goal demotion reasoning.

Crucially, the same four-step sequence handles any failure combination. If two tools fail simultaneously, both get infinite-weight edges in Step 2, and Step 3 finds the best path avoiding both. If three tools fail, same algorithm, same steps. The time complexity does not change with the number of simultaneous failures — Dijkstra runs once regardless. This is the fundamental advantage over state machines, which require a separate pre-coded edge for each failure combination.

### 2.4 Self-Healing Loop: Automatic Recovery Without LLM

The three layers form a self-healing loop. When a tool fails during execution, the Tool Health Monitor immediately fires with high priority (0.99), winning the priority competition. The orchestrator receives this signal, marks the failed tool's graph edges as infinite cost, and re-invokes Dijkstra. The algorithm finds the next cheapest path and execution continues without interruption. The entire recovery cycle is sub-millisecond — negligible compared to tool API latency — and requires zero LLM calls.

Figure 5 illustrates this loop step by step, and contrasts it with the ReAct/Agent SDK approach where every tool failure requires 3-5 LLM reasoning calls to discover and invoke an alternative. The LLM is only consulted when self-healing fails: when Dijkstra cannot find ANY path to the goal because all alternatives in a required stage are exhausted. In that case, the situation genuinely requires reasoning about how to degrade the goal.

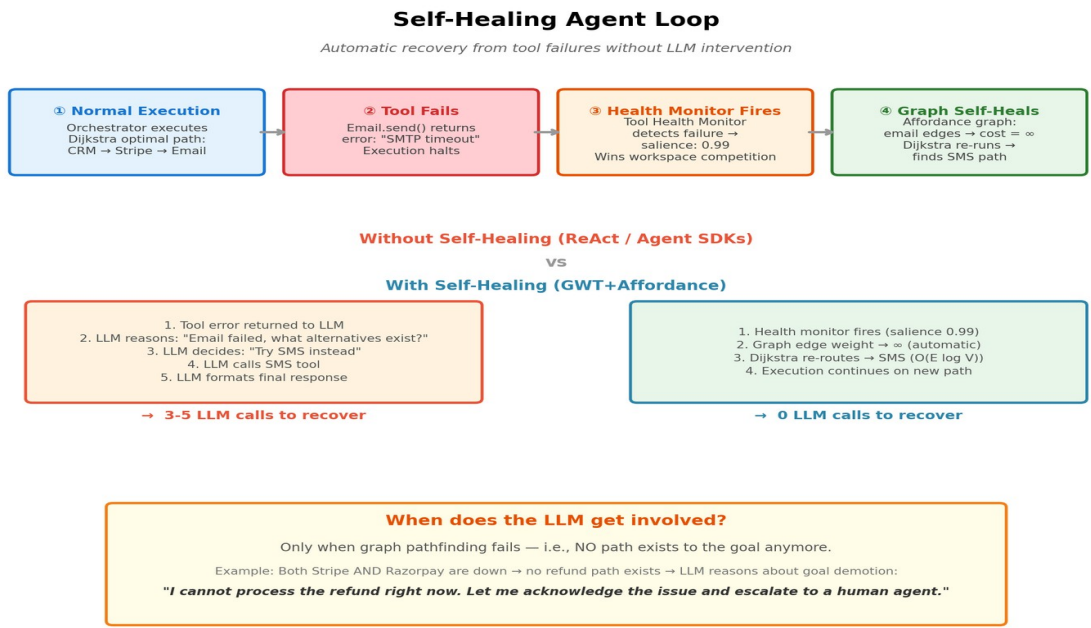

*Figure 5: Self-healing agent loop. Tool failure triggers automatic graph rerouting (0 LLM calls). The LLM is only consulted when no path to the goal exists.*

### 2.5 Runtime Graph Calibration

The benchmark results in this paper use static edge weights (cost=1 for primary tools, cost=2 for backups) that switch to infinity on hard failure. This is sufficient to demonstrate the architecture's structural properties. However, a production deployment requires continuously calibrated weights that reflect real-time tool conditions. This section proposes a calibration mechanism for production deployment. Note that existing production systems (Envoy, Istio) use layered, separated mechanisms — primarily binary health checking (with degraded state support since ~2019), independent circuit breakers, and configurable load balancing (weighted least-request, consistent hashing, locality-aware routing) — rather than a unified composite weight. The composite weight function below is a novel design contribution that unifies these separate concerns into a single pathfinding-compatible representation.

### 2.5.1 Weight Composition

In production, each edge weight is not a static number but a composite function of live telemetry:

$$W(tool) = base\_cost \times latency(t) \times reliability(t) \times rate\_limit(t) \times availability(t)$$

*where t = current time, and each factor is continuously updated from live telemetry*

| Component | Source | Update Trigger | Range |
|---|---|---|---|
| base_cost | Per-transaction fees, SLA tier pricing | Configuration change | 0.5 – 5.0 (static) |
| latency(t) | Rolling average of last N response times | Every API response | 0.5 – 10.0 (continuous) |
| reliability(t) | Error rate over sliding window (15 min / 100 calls) | Every API response | 1.0 – 50.0 (continuous) |
| rate_limit(t) | Remaining API quota / total quota | Every API call | 1.0 – ∞ (spikes near limit) |
| availability(t) | Circuit breaker state (CLOSED / OPEN / HALF-OPEN) | Health check or failure event | 1.0 or ∞ (binary) |

*Weight function components. Each factor is derived from a different telemetry source and updated at a different frequency. The composite weight gives Dijkstra a continuously calibrated view of the tool landscape.*

The base_cost reflects the inherent expense of the tool — Razorpay may charge lower transaction fees than Stripe. The latency factor is derived from a rolling average of recent response times: if Stripe normally responds in 200ms but is currently averaging 800ms, its weight increases to reflect degraded performance. The reliability factor penalizes tools with elevated error rates over a sliding window. The rate_limit factor increases sharply as the tool approaches its API quota — at 95% of the hourly limit, the weight might double; at 100%, it goes to infinity. The availability factor is the binary circuit breaker: 1.0 when healthy, infinity when tripped open.

This composite weight means Dijkstra does not merely route around failures — it continuously optimizes for the cheapest, fastest, most reliable path given current conditions. A tool experiencing latency spikes is naturally deprioritized without any explicit rule. A tool approaching its rate limit is avoided before it actually hits the limit. Hard failure is just the extreme case of continuous degradation.

### 2.5.2 Calibration Sources and Update Frequency

Edge weights are updated from two sources operating at different frequencies. Reactive updates occur on every API call: when the orchestrator calls a tool, the response time and success/failure status are fed back to the health monitor, which recalculates the tool's weight components. This means every request leaves the graph slightly more informed than the last. The cost is negligible — updating a rolling average and recalculating a weight is a handful of arithmetic operations.

Proactive updates occur on a configurable interval (typically 5-30 seconds) via background health checks. The health monitor pings each tool with a lightweight probe (HTTP HEAD or a minimal API call) and updates latency and availability scores. This serves two purposes: it detects failures before any customer-facing request encounters them, and it detects recovery after a failure — enabling the graph to restore a tool's weight before any request needs it. This is functionally identical to Kubernetes liveness probes and Envoy health check intervals.

### 2.5.3 Circuit Breaker States as Weight Transitions

The binary failure model described in Section 2.3 (healthy → infinite weight) is a special case of the well-established circuit breaker pattern (Nygard, 2007). In production, the tool graph implements the full three-state circuit breaker per tool node: CLOSED (normal operation, weight based on live telemetry), OPEN (tool confirmed down, weight set to infinity, Dijkstra routes around it), and HALF-OPEN (recovery suspected, one probe request sent, weight set high but not infinite). When the half-open probe succeeds, the circuit transitions back to CLOSED and the weight gradually returns to its telemetry-based value. When it fails, the circuit returns to OPEN.

This state machine prevents two pathological behaviors. First, it prevents flapping: if a tool oscillates between up and down, the circuit breaker holds it open for a configurable cooldown before retrying, avoiding repeated failures. Second, it ensures graceful recovery: rather than instantly restoring a tool to full weight after a single successful probe, the weight decreases gradually over several successful calls, preventing a recently-recovered tool from being overwhelmed with traffic.

### 2.5.4 Fresh Pathfinding Per Request

A critical design decision is that Dijkstra runs fresh on every incoming request rather than caching paths. Because the graph is small (typically 5-50 nodes for a single agent domain) and Dijkstra is $O((V+E) \log V)$, the computation cost is negligible relative to tool and network latency. The benefit is that every request gets the optimal path given the latest telemetry. There is no stale routing. If a tool's latency spiked 50 milliseconds ago, the very next request will route around it. This is the same design principle behind Envoy's per-request load balancing decisions, applied at the semantic tool orchestration layer.

To summarize: the tool graph is not a static structure that handles failures. It is a continuously self-calibrating routing engine where every API response refines the model, every health check updates availability, and every request receives a freshly optimized path. Hard failure and binary circuit breaking are simply the extreme end of this continuous calibration spectrum. The architecture we benchmark in this paper validates the structural properties (correct rerouting, goal demotion, zero silent failures) using static weights. The calibration layer described here extends these properties to the continuous, noisy conditions of real-world API environments. We emphasize that Section 2.5 describes proposed production design, not validated implementation. The benchmarks in this paper use static weights; the composite weight function, circuit breaker integration, and per-request recalibration described above are architectural recommendations derived from established infrastructure patterns, not empirically tested claims.

## 2.6 LLM as Last Resort: Genuine Reasoning Only

The LLM is invoked only when no graph path exists for the current goal, meaning all functional alternatives have been exhausted. The orchestrator asks the LLM to reason about goal demotion: "The original goal was refund, but all payment providers are down. What is the best fallback?" This is genuine reasoning that cannot be pre-coded, because the space of possible degraded goals depends on context.

## 3. Multi-Domain Evaluation

A key question for any agent architecture is whether it generalizes across different tool orchestration patterns, or whether it only works for one graph shape. We evaluate across three domains chosen specifically because their tool graphs have structurally distinct topologies.

Before examining per-domain results, consider how the four approaches handle the same failure. In Scenario S5, email dies mid-task after a refund has already been processed:

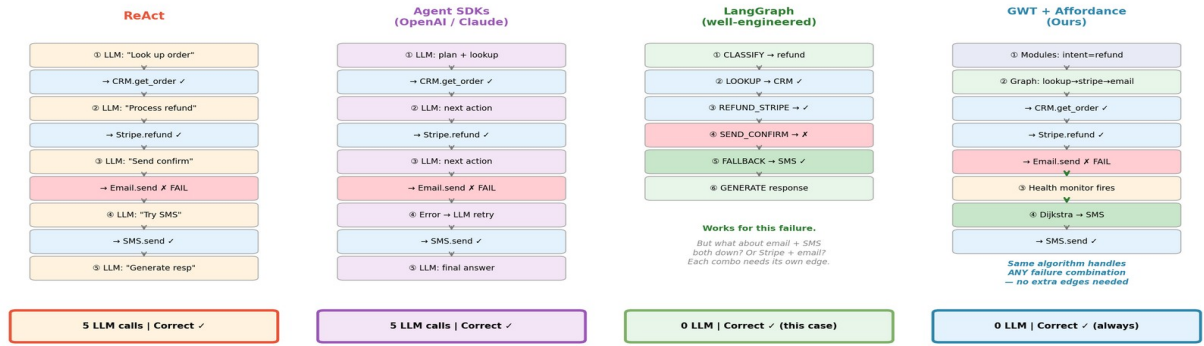

*Figure 6: Same failure (email dies mid-task) handled four ways. ReAct and Agent SDKs both need 5 LLM calls. A well-engineered LangGraph with an SMS fallback edge handles this case correctly — but needs a separate edge for every failure combination. Self-Healing Router reroutes via Dijkstra with 0 LLM calls and handles any combination.*

Note that a well-engineered LangGraph can handle this specific failure — by pre-coding an SMS fallback edge from SEND_CONFIRMATION. This is fair: good teams do anticipate common failures. The question is what happens with simultaneous failures. If email AND SMS are both down, that requires yet another edge. If Stripe AND email fail in the same request, another. For N tools with K failure modes each, the possible simultaneous-failure states number K^N. In our travel booking domain with 8 tools, even with just 2 failure modes (up/down), that is 256 combinations. Graph pathfinding handles all 256 with the same Dijkstra call — no additional engineering required.

Production Agent SDKs from OpenAI and Anthropic produce the same correct result as ReAct but with better tooling: structured error callbacks, retry policies, and guardrails. However, they inherit the same LLM-per-decision cost because every tool failure is routed back to the LLM for reasoning about what to do next.

### 3.1 Domain 1: Customer Support — Linear Pipeline

The customer support domain uses a linear pipeline topology: lookup order → process refund → send confirmation. At each stage, alternative providers exist (Stripe/Razorpay for payment, Email/SMS for notification). This tests the most basic form of graph rerouting: when one tool fails, can Dijkstra find the alternative at the same stage?

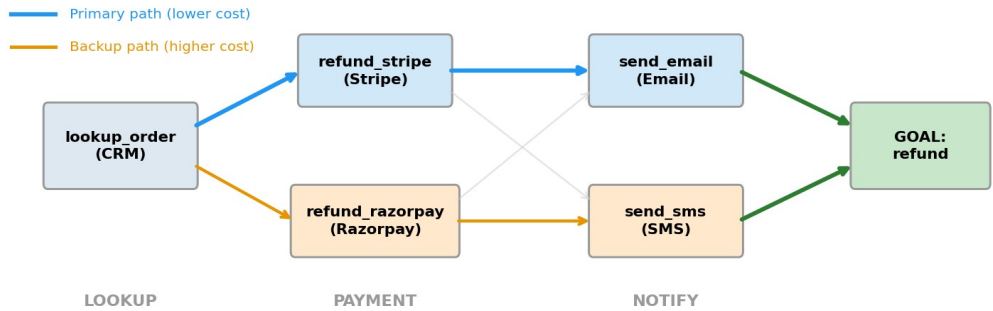

*Figure 7: Customer Support graph topology — linear pipeline with alternative providers at each stage.*

Seven scenarios test progressive failure complexity: happy path (S1), functional substitution when Stripe fails (S2), goal demotion when all payment providers fail (S3), risk interruption for high-value refunds (S4), novel mid-task failure when email dies after refund succeeds (S5), cascading notification failure when both email and SMS are down (S6), and triple simultaneous failure — Stripe + Email + SMS all down (S7).

| Scenario | SH LLM | ReAct LLM | LG LLM | SH Tools | ReAct Tools | LG Silent Fail |
|---|---|---|---|---|---|---|
| S1: Happy Path | 0 | 4 | 0 | 3 | 3 | No |
| S2: Stripe Down | 0 | 5 | 0 | 4 | 4 | No |
| S3: All Payment Down | 1 | 7 | 0 | 4 | 5 | No |
| S4: Risk ($15k) | 1 | 4 | 0 | 3 | 2 | No |
| S5: Email Dies | 0 | 5 | 0 | 4 | 4 | No |
| S6: Both Notif Down | 2 | 8 | 0 | 4 | 4 | **Yes** |
| S7: Triple Failure | 2 | 9 | 0 | 5 | 5 | **Yes** |

*Table 1: Customer Support results (7 scenarios). Well-engineered LangGraph handles S5 via SMS fallback. Fails silently on S6 (both notification channels down) and S7 (triple failure: Stripe + Email + SMS).*

### 3.2 Domain 2: Travel Booking — Dependency DAG

The travel booking domain uses a dependency DAG topology: flights → hotels (depends on flight) → car rental → confirmation. Unlike the linear pipeline, failures here can cascade downstream: if the flight API fails, the entire downstream chain (hotel, car) must be replanned through alternative providers. This tests whether Dijkstra can handle multi-level rerouting through a dependency graph.

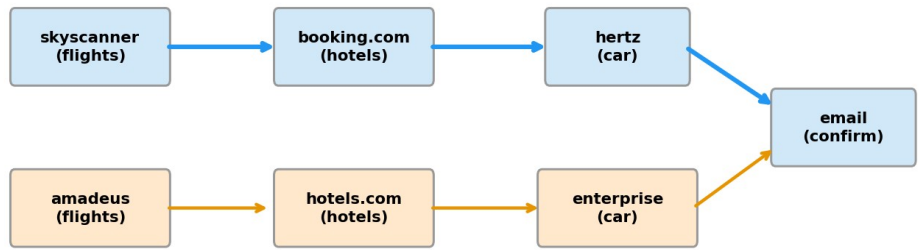

*Figure 8: Travel Booking graph topology — dependency DAG where each stage depends on the prior stage completing.*

Six scenarios test: happy path (T1), flight API failure with downstream chain intact (T2), cascading reroute through both flight and hotel backups (T3), budget risk triggering priority competition (T4), complete accommodation failure requiring LLM goal demotion to flight-only (T5), and triple simultaneous failure across flight, hotel, and confirmation (T6).

| Scenario | SH LLM | ReAct LLM | LG LLM | SH Tools | SH Recovery | ReAct Tools |
|---|---|---|---|---|---|---|
| T1: Happy Path | 0 | 5 | 0 | 4 | 0 | 4 |
| T2: Flight API Down | 0 | 6 | 0 | 5 | 1 | 5 |
| T3: Cascading | 0 | 7 | 0 | 5 | 2 | 6 |
| T4: Budget Risk | 1 | 3 | 0 | 2 | 0 | 0 |
| T5: Hotel+Car Down | 1 | 5 | 0 | 5 | 2 | 6 |
| T6: Triple Failure | 0 | 8 | 0 | 6 | 2 | 6 |

*Table 2: Travel Booking results (6 scenarios). SH-Router handles cascading reroutes (T2, T3, T6) with 0 LLM calls via Dijkstra. LangGraph fails silently on T6 (triple simultaneous failure).*

## 3.3 Domain 3: Content Moderation — Parallel Fan-Out

The content moderation domain uses a parallel fan-out topology: three classifiers (image, text, user history) run independently, feeding through optional secondary checks (toxicity API, spam filter) to a terminal action queue. Unlike the sequential topologies above, failures here cause partial degradation rather than path blockage — if the image classifier fails, the text analyzer and user history can still reach the action queue.

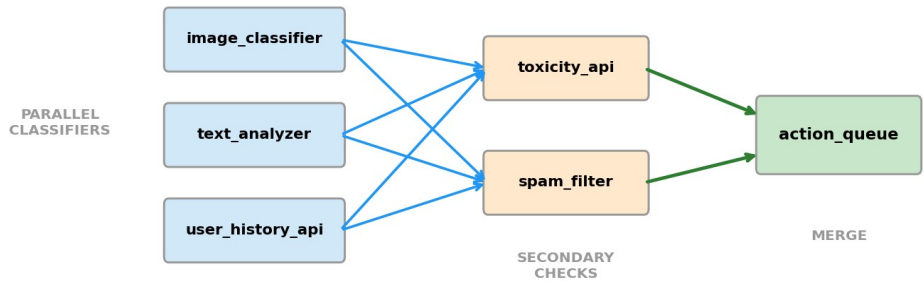

Figure 9: Content Moderation graph topology — parallel fan-out where multiple classifiers feed into an action queue.

Six scenarios test: happy path (M1), single classifier failure with graceful degradation (M2), toxicity risk signal outbidding normal intent (M3), multiple classifiers down requiring hold-for-review (M4), cascading classifier failure (M5), and extreme combinatorial failure with 4 of 5 classifiers down simultaneously (M6).

| Scenario | SH LLM | ReAct LLM | LG LLM | SH Tools | ReAct Tools | LG Classifiers Lost |
|---|---|---|---|---|---|---|
| M1: Happy Path | 0 | 6 | 0 | 2 | 6 | 0 |
| M2: Image Down | 0 | 7 | 0 | 2 | 6 | 1 |
| M3: Toxicity Risk | 1 | 6 | 0 | 2 | 6 | 0 |
| M4: Multi Down | 0 | 9 | 0 | 2 | 5 | **3** |
| M5: Cascading | 0 | 9 | 0 | 2 | 5 | **3** |
| M6: 4/5 Down | 0 | 10 | 0 | 2 | 5 | **4** |

Table 3: Content Moderation results (6 scenarios). Well-engineered LangGraph now holds for review when 3+ classifiers fail (M4-M6). All three architectures produce defensible outcomes in this domain.

## 4. Cross-Domain Results

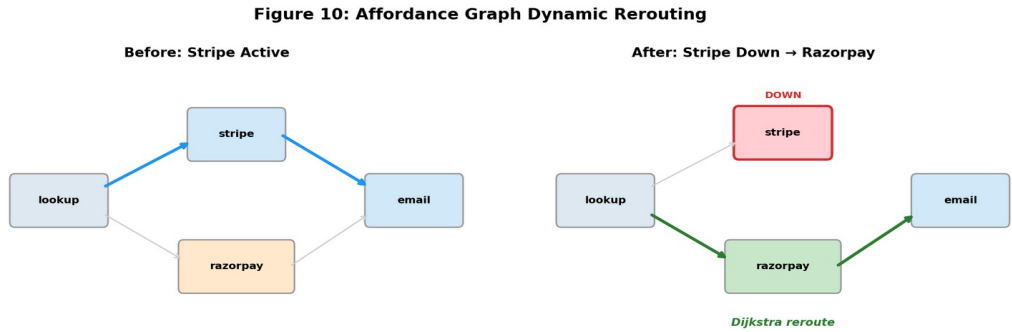

*Figure 10: Tool graph dynamic rerouting — Dijkstra automatically finds alternative paths when tools fail.*

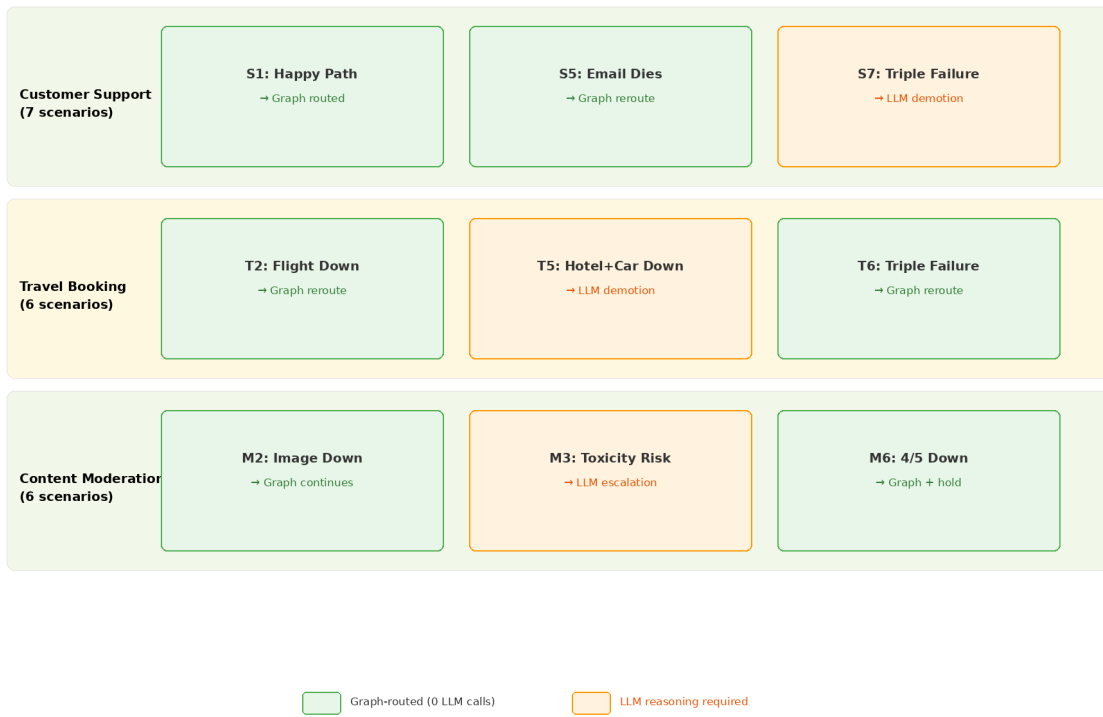

*Figure 11: Representative scenario flows across all three domains. Green = graph-routed (0 LLM); amber = LLM reasoning required.*

### 4.1 Aggregate Results

| Architecture | Correct | LLM Reasoning | Tools | Recovery | Silent Fail |
|---|---|---|---|---|---|
| **Self-Healing Router** | 19/19 | 9 | 66 | 13 | 0 |
| **ReAct** | 19/19 | 123 | 93 | 0 | 0 |
| **LangGraph (well-eng.)** | 16/19 | 0 | 87 | 24 | 3 |

*Table 4: Grand total across 19 scenarios and 3 domains. LangGraph baseline is well-engineered with single-failure fallback edges.*

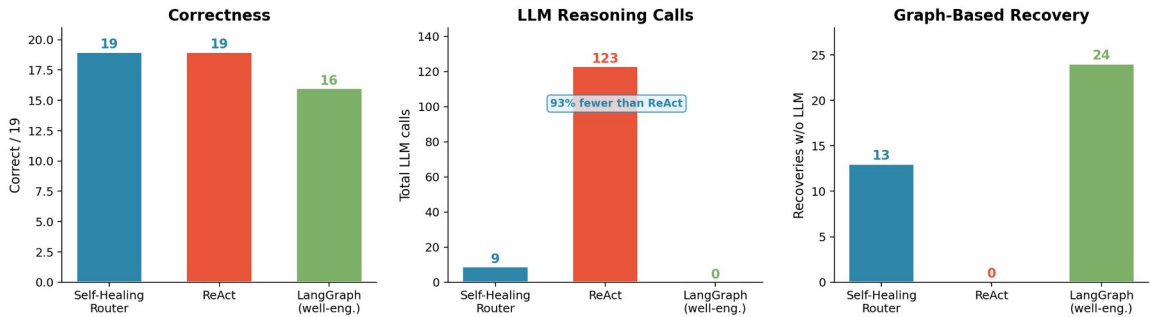

*Figure 12: Cross-domain comparison — correctness, LLM cost, and graph-based recovery.*

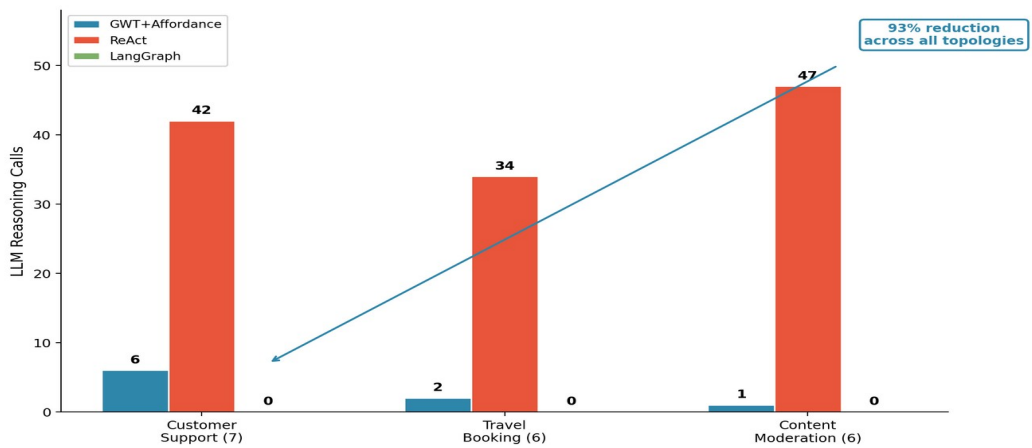

*Figure 13: LLM reasoning calls by domain. Self-Healing Router uses 93% fewer calls than ReAct across all topologies.*

The results demonstrate four key findings. First, Self-Healing Router achieves 100% correctness (19/19) while even a well-engineered LangGraph with single-failure fallback edges fails silently on 3 of 19 scenarios — always in cases involving combinatorial failures where multiple tools fail simultaneously. Second, Self-Healing Router uses 93% fewer LLM reasoning calls than ReAct (9 vs 123), because graph-based pathfinding handles routine decisions. Third, 13 failure-recovery events were handled entirely by Dijkstra rerouting, with zero LLM involvement. Fourth, these properties hold across three structurally distinct graph topologies — linear pipeline, dependency DAG, and parallel fan-out — demonstrating architectural generalization. The LangGraph baseline is generous: it includes explicit SMS fallback for email failures and threshold-based hold-for-review in content moderation. The failures it encounters are genuinely hard — they require reasoning about compound failure states that no static edge set can anticipate without combinatorial explosion. Note: these results validate control-flow properties (reroute-vs-escalate behavior, call-count reduction, observability guarantees), not semantic correctness under real tool outputs with partial failures or malformed responses.

## 4.2 Recovery Complexity Analysis

Table 5 formalizes the recovery time complexity for each architecture across different failure scenarios. The key insight is that Self-Healing Router recovery complexity is invariant to the number of simultaneous failures — Dijkstra runs once on the updated graph regardless of how many edges were set to infinity.

| Failure Type | ReAct / Agent SDKs | LangGraph (well-eng.) | Self-Healing Router |
|---|---|---|---|
| Single tool failure | $O(k \times T\_LLM)$ k = 3-5 reasoning calls | $O(1)$ — single edge traversal (if edge exists) | $O((V+E) \log V)$ sub-millisecond |
| K simultaneous failures | $O(K \times k \times T\_LLM)$ linear in failure count | $O(\infty)$ — silent failure (no compound edges) | $O((V+E) \log V)$ same as single failure |
| All alternatives exhausted | $O(k \times T\_LLM)$ LLM reasons about failure | $O(\infty)$ — silent failure (no edge for this case) | $O((V+E) \log V) + O(T\_LLM)$ graph detects, LLM demotes |
| Recovery detection | Always (LLM reasons about every failure) | Only for pre-coded failures; silent otherwise | Always (Dijkstra returns path or null — binary) |

*Table 5: Recovery complexity by failure type. $T\_LLM \approx$ 300-500ms per inference call. V = tool nodes (typically 5-50), E = edges (typically 10-100). Self-Healing Router complexity does not scale with the number of simultaneous failures.*

## 4.3 Risk Exposure at Scale

Table 6 projects the operational impact of each architecture at different deployment scales, using a 5% tool failure rate as a sensitivity baseline; results scale linearly with the actual failure rate.

| Scale (tasks/day) | Recovery events/day | ReAct: cumulative recovery time | ReAct: LLM calls for recovery | LangGraph: silent failures/day | SH-Router: recovery time |
|---|---|---|---|---|---|
| 10,000 | 500 | ~17 minutes | ~2,000 | 10-25 | < 1 second |
| 100,000 | 5,000 | ~170 minutes | ~20,000 | 100-250 | < 5 seconds |
| 1,000,000 | 50,000 | ~28 hours | ~200,000 | 1,000-2,500 | < 50 seconds |

*Table 6: Operational risk exposure at scale. Uses 5% tool failure rate as a sensitivity point, ~4 LLM calls per recovery event (ReAct), ~2 seconds per recovery, and an estimated 2-5% compound-failure rate among failure events (LangGraph silent failures — 3/19 scenarios in our benchmark involve compound failures, but single-tool failures dominate in production). Silent-failure exposure scales linearly with task volume and compound-failure rate; the specific values shown illustrate relative magnitude across architectures rather than precise production predictions.*

The silent failure column warrants emphasis. Slow recovery (ReAct) is visible and measurable — operations teams can track recovery latency, set alerts, and optimize. Silent failure (LangGraph compound cases) is invisible until a customer complains or an end-to-end audit catches the gap. At enterprise scale, the potential for undetected incidents — scaling linearly with task volume and compound-failure rate — represents significant operational and compliance risk, particularly in regulated industries where every customer interaction must be fully documented.

*Graph-Based Self-Healing Tool Routing for Cost-Efficient LLM Agents*

*Neeraj Bholani, 2026*

5. Discussion

5.1 Why Three Topologies Matter

The three domains were chosen because they produce genuinely different graph structures that test different aspects of the architecture. The linear pipeline (customer support) tests single-stage substitution. The dependency DAG (travel booking) tests cascading rerouting across multiple dependent stages. The parallel fan-out (content moderation) tests graceful degradation when some parallel paths fail. A single-domain evaluation could not distinguish an architecture that works for one graph shape from one that generalizes.

5.2 Relationship to Production Agent SDKs

Production agent SDKs — including OpenAI Agents SDK and Anthropic's Claude Agent SDK — represent the current state of the art for building tool-using agents. Both provide structured error handling, retry logic, and function-calling interfaces that significantly improve developer experience over raw ReAct loops. However, both inherit the same fundamental cost structure: when a tool fails, the error is passed back to the LLM for reasoning about what to do next. Our benchmark shows this pattern requiring 123 LLM reasoning calls across 19 scenarios. Our architecture proposes a complementary pre-filter layer: the tool graph handles routine rerouting (93% of the decisions that would otherwise require LLM inference) before the LLM is ever consulted, while preserving the LLM as the fallback for genuinely novel situations. This is not a replacement for production SDKs — it is a layer that could sit within any agent framework to reduce the volume of decisions that reach the LLM.

5.3 Extension to Multi-Agent Systems

This paper evaluates a single-agent architecture. We note briefly that the health monitoring signals generated by the Tool Health Monitor could be shared across agent instances operating on the same tool ecosystem. When Agent A detects that Stripe is down, it could broadcast this signal to Agent B's workspace, enabling Agent B to pre-emptively reroute before encountering the failure itself. The tool graph could be shared or federated, with each agent maintaining local weights but consuming global health signals. We leave the coordination protocol, conflict resolution for concurrent graph updates, and distributed weight consistency as future work.

5.4 Honest Limitations

We identify several limitations that should be addressed in future work:

Mock environment. All 19 scenarios use mock tools and a deterministic mock LLM. This validates structural properties — that Dijkstra correctly reroutes, that priority competition correctly prioritizes, that goal demotion triggers at the right time — but does not validate real-world API latency, error distributions, or LLM accuracy under production conditions.

Graph construction cost. The tool graph must be manually specified for each domain. We do not address the cold-start problem of automatically constructing the graph from tool documentation or usage patterns, though this is a promising direction for future work.

Module accuracy. In production, intent classifiers and risk detectors would be learned models with non-zero error rates. Our evaluation assumes perfect module accuracy, which overstates the architecture's reliability. The priority competition mechanism provides some robustness (a misclassified intent with low priority will be outbid by a correct risk signal), but systematic evaluation of module error propagation is needed.

Limited failure modes. Our scenarios test tool outages and cascading failures, but do not test partial failures (tool returns incorrect data), latency degradation, or adversarial inputs. Real-world robustness would require more diverse failure injection.

## 5.5 Production Readiness: Recovery Time, Task Integrity, and Failure Observability

Agent architectures are typically evaluated on correctness and cost. But production deployments care about three additional properties that current frameworks largely ignore: how fast can you recover from a tool failure, how much completed work is preserved during recovery, and does the system know when it has failed? We examine each through the lens of our benchmark results.

### 5.5.1 Recovery Time

Recovery time — analogous to Recovery Time Objective (RTO) in infrastructure engineering — measures the elapsed time between a tool failure and the agent resuming correct execution. The three architectures exhibit fundamentally different recovery profiles:

ReAct and Agent SDKs require LLM inference for every recovery step. When email fails in Scenario S5, the LLM must (1) recognize the failure, (2) reason about alternatives, (3) select SMS, and (4) format the new call — approximately 3-5 LLM inference calls at ~300-500ms each. Total recovery time: 1-2.5 seconds per failure event. Critically, this scales linearly: in Scenario T6 with three simultaneous failures, ReAct requires ~8 LLM calls totaling 3-4 seconds of recovery time.

LangGraph with pre-coded fallback edges recovers in approximately 10 milliseconds for anticipated single failures — the time to traverse one state transition. However, for unanticipated compound failures (S6, S7, T6), recovery time is effectively infinite: the system does not recover at all, proceeding to completion with silently dropped steps.

Self-Healing Router recovers in sub-millisecond time regardless of the number of simultaneous failures. The sequence — health monitor fires, edge weights set to infinity, Dijkstra re-runs — involves only in-memory operations on a small graph. For our benchmark graphs (5-8 nodes, 10-15 edges), Dijkstra completes in sub-millisecond time — negligible compared to the 50-2000 ms of a typical tool API call. This does not change whether one tool fails or five fail simultaneously: Dijkstra runs once on the updated graph.

At production scale, the difference compounds. Consider an enterprise running 10,000 agent tasks per day with an assumed 5% tool failure rate (500 recovery events daily at this sensitivity point). ReAct accumulates approximately 17 minutes of recovery time per day and consumes ~2,000 additional LLM calls purely for recovery. Self-Healing Router accumulates 0.5 seconds total. LangGraph accumulates near-zero for anticipated failures, but the potential for undetected partial completions scales with the compound-failure rate — a risk that grows with task volume and tool-chain complexity.

### 5.5.2 Task Integrity

Task integrity — analogous to Recovery Point Objective (RPO) — measures how much completed work is preserved when recovery occurs. In agent workflows, this translates to a concrete question: if a tool fails mid-task, do the results of prior successful steps survive?

Consider Scenario S5: the CRM lookup succeeds, the Stripe refund succeeds, then email confirmation fails. The refund is already processed and cannot be undone. The question is whether the customer is ever notified.

Self-Healing Router preserves all prior work and completes the remaining steps via alternative paths. The refund stays processed; the notification reroutes to SMS. Task integrity: 100% — no completed work lost, no pending work dropped.

ReAct achieves the same 100% task integrity, but at higher cost (additional LLM calls to discover the alternative path).

LangGraph preserves prior work (the refund stays processed) but may silently drop remaining steps. In S6, the customer receives a refund but is never notified through any channel. In T6, flights and hotels are booked but the confirmation email is never sent. The customer has a valid booking they do not know about. Task integrity is partial: prior work preserved, but the task is incomplete in ways the system does not report.

The partial-integrity case is particularly dangerous in regulated industries. A financial agent that processes a trade but fails to send the confirmation creates a compliance gap. A healthcare agent that schedules a procedure but drops the patient notification creates a safety risk. In both cases, the system reports success.

### 5.5.3 Failure Observability

The most operationally significant property may be failure observability: does the system know when it has degraded? This determines whether engineering teams can detect, alert on, and remediate failures before they impact end users.

Self-Healing Router has binary failure states. At every decision point, Dijkstra either returns a valid path (the system continues normally on an alternative route) or returns null (no path exists, and the system explicitly escalates to LLM reasoning for goal demotion). There is no intermediate state. Every reroute is logged. Every escalation is explicit. An operations team can monitor the ratio of primary-path vs rerouted-path executions, set alerts when reroute rates exceed thresholds, and track which tools are most frequently bypassed.

ReAct and Agent SDKs also have good observability, because the LLM explicitly reasons about every failure. The recovery reasoning is visible in the conversation trace. The cost is that this observability comes at LLM inference cost.

LangGraph has an observability gap for compound failures. When the system encounters a failure combination it was not designed for, it does not crash, does not raise an error, and does not log a warning — it simply proceeds without the dropped step. The state machine completes successfully. The output looks normal. The only way to detect the failure is end-to-end auditing of task completion, which requires comparing what the agent did against what it should have done. In our benchmark, 3 of 19 scenarios (16%) produce this silent failure mode.

For production operations teams, this means: Self-Healing Router failures are always visible (either as reroutes or as LLM escalations). ReAct failures are always visible (as LLM reasoning traces). LangGraph compound failures are invisible until a customer complains or an audit catches the gap. In SLA-governed environments, invisible failures are strictly worse than slow failures.

## 5.6 Related Work

Our architecture draws on and extends four distinct bodies of work: LLM agent frameworks, service mesh infrastructure, workflow orchestration engines, and cognitive architecture theory (as intellectual lineage). Notably, no existing system combines the capabilities from all four. We position our contribution by examining what each provides and what it lacks.

### 5.6.1 LLM Agent Frameworks

ReAct (Yao et al., 2023) established the dominant agent loop: the LLM observes, reasons, and acts on every step. This guarantees correctness — the LLM can handle any failure by reasoning about alternatives — but incurs O(n) LLM calls per task. Our benchmark confirms this cost: 123 LLM reasoning calls across 19 scenarios. Production SDKs from OpenAI (Agents SDK) and Anthropic (Claude Agent SDK) wrap this pattern in structured error handling, retry logic, and guardrails, but inherit the same per-decision LLM cost because every tool failure routes back to the model.

Toolformer (Schick et al., 2023) taught language models to decide when and how to call tools, embedding tool use into the generation process itself. This reduces the explicit reasoning overhead but still relies on the LLM for tool selection and error handling.

LATS (Zhou et al., 2023; ICML 2024) introduced tree search over action spaces, using LLM-guided Monte Carlo Tree Search to explore tool-use strategies. This improves decision quality for complex tasks but amplifies LLM cost — each branch of the tree requires LLM evaluation.

LangGraph and LangChain represent the state-machine approach: developers pre-define nodes (tools) and edges (transitions), creating deterministic workflows that require zero LLM calls for anticipated paths. Our evaluation provides LangGraph with a generous baseline including explicit fallback edges for single failures. The architecture handles anticipated failures well but encounters silent failures when compound failure combinations arise that were not pre-coded.

Voyager (Wang et al., 2023) introduced a skill library where the agent discovers and stores reusable action sequences, executable code, and iterative self-verification. This is complementary to our approach — discovered skills could become nodes in the tool graph, with Dijkstra handling routing between them.

ControlLLM (Liu et al., ECCV 2024) is the closest prior art to our graph-based routing mechanism. ControlLLM constructs a tool graph where nodes are tools and edges encode parameter dependencies, then uses depth-first search to find solution paths. Our approach differs in three ways: we use cost-weighted Dijkstra rather than unweighted DFS, enabling preference ordering among alternatives; our graph edges are dynamically reweighted based on real-time health telemetry; and our architecture implements self-healing recovery — when a tool fails mid-execution, Dijkstra automatically re-routes without any LLM involvement, which ControlLLM does not address. The key distinction is that ControlLLM uses graphs for task decomposition in multimodal pipelines, while we use them for fault-tolerant tool routing with automatic recovery.

Two other graph-based tool orchestration systems deserve comparison. ToolNet (Liu et al., 2024) organizes tools into a directed weighted graph and has the LLM navigate iteratively through tool nodes — the most structurally similar prior system. However, ToolNet uses LLM-driven navigation rather than algorithmic pathfinding (no Dijkstra), has no parallel health monitoring, and lacks automatic recovery on failure. NaviAgent (Qin et al., 2025; v2 October 2025) builds tool dependency graphs with bilevel planning and features a continuously evolving Tool World Navigation Model updated via execution feedback, with adaptive path recombination when tools fail — the closest to combining graph planning with failure recovery. However, NaviAgent's recovery is integrated into its LLM-driven planning loop (using heuristic search tested against Alpha-Beta pruning), not a separate deterministic mechanism; and it lacks a dedicated health monitoring daemon that fires independently of the planning cycle. APBDA (Mirchev et al., 2025) is the closest prior work using Dijkstra for AI agent routing: it applies a priority-based Dijkstra variant with a multi-factor cost function and reinforcement-learning-based weight adaptation. However, APBDA routes tasks between AI agents in a network, not tools within a single agent's workflow; it uses RL for weight adaptation rather than binary health signals from runtime telemetry; and it has no self-healing or failure recovery mechanism. Our architecture differs from all three by using Dijkstra as a deterministic pathfinding algorithm driven by live health telemetry for intra-agent tool routing with automatic recovery, rather than inter-agent task dispatch or LLM-driven navigation.

Two recent works address the specific challenge of agent tool failure recovery. PALADIN (Vuddanti et al., 2025) introduces failure injection training across seven canonical error classes, achieving 89.68% recovery rates through a dual mechanism: LoRA fine-tuning on 50K+ failure trajectories and inference-time taxonomic retrieval from 55+ curated failure exemplars aligned with a ToolScan taxonomy. Where PALADIN teaches the LLM to recover through training data and retrieval, our architecture recovers through structure — Dijkstra rerouting requires no training data and handles novel failure combinations that were never seen during development. The approaches are complementary: PALADIN could improve LLM-based goal demotion quality in the cases where our graph-based recovery is exhausted. SHIELDA (Zhou et al., 2025) proposes a reactive runtime exception handling framework with a taxonomy of 36 exception types across 12 agent artifacts and a triadic handler pattern (Local Handling, Flow Control, State Recovery). SHIELDA classifies and handles exceptions as they occur at runtime, while our architecture prevents routing-level failures by construction — Dijkstra never routes through a failed tool because its edges are already at infinite weight.

Table 7 summarizes the key architectural differences between graph-based tool orchestration systems. Our focus is runtime fault tolerance with deterministic recovery and observability, not large-scale tool discovery or toolchain planning.

| System | Primary Aim | Recovery w/o LLM | Deterministic Bounded-Time | Dynamic Edge Weights | Observability Guarantee |
|---|---|---|---|---|---|
| ControlLLM | Tool planning (multimodal) | No | No (DFS) | No (static) | No |
| ToolNet | Tool discovery (large libraries) | No | No (LLM nav) | Yes (staleness) | No |
| NaviAgent | Large-scale orchestration | Partial (path recomb.) | No (search) | Yes (adaptive) | No |
| APBDA | Agent routing (inter-agent) | No | Yes (Dijkstra + RL) | Yes (RL-based) | No |
| PALADIN | Failure recovery (training-based) | No (LoRA) | No (learned) | N/A | No |
| SHIELDA | Exception handling | Partial (post-hoc) | No (LLM classify) | N/A | No |
| Self-Healing Router (Ours) | Fault-tolerant routing | Yes (Dijkstra) | Yes (O((V+E) log V)) | Yes (live telemetry) | Yes (path or null) |

Table 7. Comparison of graph-based tool orchestration systems. Our architecture is the only system combining deterministic algorithmic recovery, bounded-time complexity, live edge reweighting, and binary observability (valid path or explicit escalation).

Our architecture occupies the middle of the spectrum these frameworks define — a spectrum explicitly articulated by Chase (LangChain Blog, 2024) as ranging from hardcoded workflows to fully autonomous agents. Like LangGraph, routine decisions are handled without LLM calls. Like ReAct, novel situations trigger LLM reasoning. The contribution is the mechanism for separating the two: cost-weighted pathfinding on a dynamic graph, rather than either static edges or LLM inference.

### 5.6.2 Service Mesh and Infrastructure Patterns

Service meshes such as Istio and Envoy provide sophisticated traffic management at the network layer: health checking, circuit breaking, weighted routing, and automatic failover between service instances. The tool graph draws direct inspiration from these patterns — our tool health monitor functions like a circuit breaker, and our cost-weighted edges function like weighted routing rules.

The critical difference is semantic awareness. A service mesh routes HTTP requests between interchangeable service replicas. It does not know that Stripe and Razorpay are functionally equivalent payment processors, that a failed email notification can be rerouted to SMS, or that when all payment options are exhausted the agent should demote its goal rather than retry indefinitely. Our architecture adds this semantic layer: the tool graph encodes functional equivalence and goal-directed fallback logic, while Dijkstra provides the same efficient pathfinding that service meshes use internally.

The circuit breaker pattern (Nygard, 2007; popularized by Netflix Hystrix and subsequently Resilience4j and Polly) is perhaps the closest infrastructure analogue to our approach. A circuit breaker detects failure, trips open, and routes to a fallback — exactly what our health monitor and edge-weight update accomplish. However, circuit breakers operate on individual service calls. They do not perform cross-service pathfinding: if Service A fails and the fallback is a three-step sequence through Services B, C, and D, the circuit breaker cannot discover this path. Our tool graph can, because Dijkstra operates on the full tool graph, not on individual connections.

### 5.6.3 Workflow Orchestration Engines

Workflow engines such as Temporal, Apache Airflow, and Prefect execute DAG-structured task pipelines with retry policies, timeout handling, and conditional branching. These systems are battle-tested in production and handle many of the reliability concerns we address.

The key limitation is that workflow DAG edges are static. When a task fails, the engine retries it or follows a pre-coded fallback branch — it does not dynamically discover alternative paths through the remaining healthy tasks. Adding a new fallback requires modifying the DAG definition, redeploying, and restarting affected workflows. In contrast, our tool graph reroutes dynamically at runtime: when a tool fails, Dijkstra finds the best alternative path without any DAG redefinition.

Additionally, workflow engines do not incorporate cost-weighted routing. All paths are treated as either available or failed, with no concept of preferring cheaper or faster alternatives when multiple options exist. The tool graph's cost weights enable this preference ordering naturally.

### 5.6.4 Intellectual Lineage: Cognitive Architectures

Our parallel-compete-broadcast pattern for health monitoring has roots in cognitive architecture research, though we apply it as an engineering pattern rather than a cognitive model. Global Workspace Theory (Baars, 1988; Baars and Franklin, 2003) proposed that specialized processors compete for access to a shared workspace, with the winning signal broadcast globally. Dehaene et al. (1998) formalized this as a neuronal model of the global workspace. The LIDA architecture, originating with CMattie (Franklin and Graesser, 1999), implemented this computationally with attention codelets competing in a global workspace.

Several recent works have applied these cognitive patterns to LLM agents. CogniPair (Ye et al., 2025) implements Global Neuronal Workspace Theory for social simulation (speed dating and hiring matching). Romero et al. (2024; AAAI Fall Symposia 2023) proposed a LIDA-inspired LLM agent architecture. CoALA (Sumers et al., 2024) provides the foundational taxonomy for cognitive LLM agent architectures. The Unified Mind Model (Hu and Ying, 2025) proposes GWT as the macro-architecture for autonomous LLM agents with a concrete agent-building engine (MindOS). Our work differs from all of these in scope and intent: we extract only the parallel-compete-broadcast pattern and apply it narrowly to tool health monitoring and priority-based routing — not to general cognition or social simulation.

The concept of affordances (Gibson, 1979; Norman, 1988, originally titled The Psychology of Everyday Things) — action possibilities offered by the environment — also informs our tool graph design. In robotics, AutoGPT+P (Birr et al., RSS 2024) combines affordance-based representations with LLM planning. Our tool graph encodes a similar idea for software tools: each node represents a capability, with edge weights encoding cost and availability in the current state.

### 5.6.5 The Integration Gap

Each of these four communities has solved part of the problem. Service meshes provide health monitoring and weighted routing. Workflow engines provide DAG execution with retry logic. Agent frameworks provide goal-aware tool orchestration — ControlLLM and ToolNet use graph search for tool path planning, NaviAgent adds adaptive path recombination, and APBDA applies Dijkstra with RL-based weight adaptation for inter-agent routing. Agent resilience research provides failure recovery — PALADIN through training-based self-correction, SHIELDA through reactive runtime exception handling. No existing system, however, combines cost-weighted graph pathfinding with parallel health monitoring specifically for fault-tolerant tool orchestration with automatic recovery.

Our architecture unifies infrastructure patterns (health checks from service meshes, DAG structure from workflow engines, circuit breaking from resilience libraries) with agent patterns (goal awareness from ReAct, graph-based tool routing from ControlLLM, deterministic workflows from LangGraph). The specific integration — dynamic cost-weighted Dijkstra pathfinding driven by parallel health monitors, with LLM reserved exclusively for goal demotion — is, to our knowledge, novel. It is precisely this integration that enables the production readiness properties described in Section 5.5.

## 6. Conclusion

We have presented a self-healing routing architecture for LLM agents that uses parallel health monitors for priority-based attention allocation and cost-weighted tool graphs for Dijkstra-based routing. By separating cheap graph-based decisions from expensive LLM reasoning, the architecture achieves both correctness and cost-efficiency across three structurally distinct domains. The key insight is that most agent decisions are not novel reasoning problems but contextually variable routing problems — and routing problems have centuries of algorithmic solutions that are faster, cheaper, and more reliable than LLM inference.

Evaluated across 19 scenarios spanning linear pipeline, dependency DAG, and parallel fan-out topologies, Self-Healing Router achieves 19/19 correctness with 93% fewer control-plane LLM calls than ReAct (measuring routing and recovery decisions, not semantic reasoning) and zero silent failures versus a well-engineered LangGraph's 3 combinatorial-failure cases. Thirteen failure-recovery events were handled entirely by graph rerouting without LLM involvement. Beyond correctness and cost, the architecture provides sub-millisecond theoretical recovery time (vs 1-2.5 seconds for LLM-based recovery), 100% task integrity (vs silent step-dropping in state machines), and binary failure observability (every degradation is either a logged reroute or an explicit LLM escalation — never a silent skip). These properties — rarely discussed in agent architecture research — are precisely what production deployments in regulated industries require. The tool graph layer proposed here could serve as a complementary pre-filter within any agent framework, including production SDKs from OpenAI and Anthropic, to reduce LLM cost while guaranteeing that failures are always fast, complete, and visible.

## Acknowledgments

This paper was developed with LLM assistance (Claude, Anthropic) for drafting, editing, literature search, and reference verification. All architectural decisions, benchmark design, and analytical claims are the author's own.

Appendix A: Self-Healing Router — Core Loop Pseudocode

The following pseudocode captures the complete runtime behavior of the Self-Healing Router. The four-step recovery sequence (detect → reweight → recompute → resume) is the core technical contribution.

```
function EXECUTE_TASK(goal, tool_graph):
    path ← DIJKSTRA(tool_graph, START, goal)
    if path is NULL: return LLM_GOAL_DEMOTION(goal)
    i ← 0
    while i < len(path):
        node ← path[i]
        result ← EXECUTE_TOOL(node)
        if result is SUCCESS:
            i ← i + 1
            continue
        // FAILURE DETECTED
        // Step 1: Health monitor fires (priority 0.99)
        signal ← MONITORS.run_all(context)
        winner ← MAX(signal.priorities)
        // Step 2: Reweight failed edges
        for each edge adjacent to node:
            edge.weight ← INFINITY
        // Step 3: Recompute path from current position
        path ← DIJKSTRA(tool_graph, node, goal)
        // Step 4: Resume or escalate
        if path is NULL:
            return LLM_GOAL_DEMOTION(goal)  // only LLM call
        i ← 0  // restart on new path
    return SUCCESS(completed_steps)
```

Key properties: (1) Recovery is $O((V+E) \log V)$ regardless of K simultaneous failures — Dijkstra runs once on the updated graph. (2) The LLM is invoked if and only if Dijkstra returns null. (3) Every execution terminates in one of two observable states: SUCCESS with a logged path, or ESCALATION with an explicit LLM-generated demotion. Silent partial completion is structurally impossible.